\documentclass{article}
\usepackage{spconf,amsmath,amssymb,dsfont,epsfig,multirow,tabularx,graphicx}
\usepackage[colorinlistoftodos,prependcaption,textsize=tiny]{todonotes}

\usepackage{fancyhdr}
\begin{document}\sloppy

\def\x{{\mathbf x}}
\def\L{{\cal L}}

\title{Visual Relationship Detection without stating the obvious with knowledge distillation: making use of what is said and omitted}
\title{Improving Visual Relationship detection by predicting relevant proposals}
\title{Improving Visual Relationship detection by predicting relevant object pairs proposals}
\title{Improving Visual Relationship detection through guided object pairs proposals}
\title{Improving Visual Relationship detection through guided object pairs proposals and semantic knowledge distillation}
\title{Improving Visual Relationship detection through guided proposals and semantic knowledge distillation}
\title{Incorporating knowledge into Visual Relationship Detection models for more accurate and relevant detections}
\title{Incorporating knowledge into Visual Relationship Detection models for improved training and inference}
\title{Semantic knowledge distillation into guided Visual Relationship Detection models}
\title{Improving guided Visual Relationship detection through semantic knowledge distillation}
\title{Semantically guided Visual Relationship detection}
\title{Visual Relationship detection with guided proposals based on semantic knowledge distillation}
\title{Visual Relationship detection based on guided proposals and semantic knowledge distillation}
%

\name{Fran\c{c}ois Plesse\textsuperscript{1,2}, Alexandru Ginsca\textsuperscript{1}, Bertrand Delezoide\textsuperscript{1}, Fran\c{c}oise Pr\^{e}teux\textsuperscript{2}}
\address{\textsuperscript{1}CEA, LIST, F-91191 Gif-sur-Yvette, France\\\textsuperscript{2}CERMICS, Ecole des Ponts, Champs-sur-Marne, France \\
\{francois.plesse, alexandru.ginsca, bertrand.delezoide\}@cea.fr; francoise.preteux@enpc.fr}
\maketitle

\begin{abstract}
A thorough comprehension of image content demands a complex grasp of the interactions that may occur in the natural world. One of the key issues is to describe the visual relationships between objects.
When dealing with real world data, capturing these very diverse interactions is a difficult problem. It can be alleviated by incorporating common sense in a network. For this, we propose a framework that makes use of semantic knowledge and estimates the relevance of object pairs during both training and test phases. Extracted from precomputed models and training annotations, this information is distilled into the neural network dedicated to this task. Using this approach, we observe a significant improvement on all classes of Visual Genome, a challenging visual relationship dataset. A 68.5\% relative gain on the recall at 100 is directly related to the relevance estimate and a 32.7\% gain to the knowledge distillation.

\end{abstract}
\section{Introduction}
\label{sec:intro}
Image understanding has lately received a lot of attention. 
It has witnessed many advances thanks to important breakthroughs in object detection \cite{Ren2015,Redmon2016}, segmentation, automatic image captioning, and most recently in visual relationship prediction. Indeed, object detection is only the first step towards image understanding, as images are more than the sum of their parts and cannot be fully understood without the relationships between these objects. The dimensionality of the output space in the case of models predicting relationships is much higher than for object detection models, which increases the scalability issues of typical classification schemes. As the relationships follow a more marked long tail distribution, it becomes even more difficult to predict relationships as classes, disjointly from the visual context. In order to overcome this hurdle, most recent models \cite{lu2016visual,ramanathan_CVPR15,li_CVPR_2017,Xu2017,dai_CVPR_2017,Newell,Yu_2017_ICCV,liang_CVPR2017,Zhu,Peyre_2017_ICCV} separately predict object and predicate classes and devise models that aim to capture statistical dependencies between the object and predicate variables.

In this work, we first aim to push the boundaries of current approaches in regards to the number of different visual relationships to be identified. We depart from the established evaluation setting, in which systems are put to test on at most 150 objects and 70 predicates \cite{lu2016visual,Xu2017}. In contrast, we investigate the performance of several approaches on a collection of 20,000 object classes and 10,000 predicates.
Existing models are not always able to completely capture such dependencies from the available data, especially in the context which we focus on. Therefore we propose a new probabilistic model that translates predicate similarities into probability densities and distills this knowledge during the training phase. This increases data efficiency and the stability of the model during the training phase, which proves useful in such contexts.

\begin{figure}[t]
\includegraphics[width=0.32\linewidth]{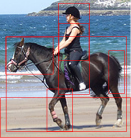}
\includegraphics[width=0.67\linewidth]{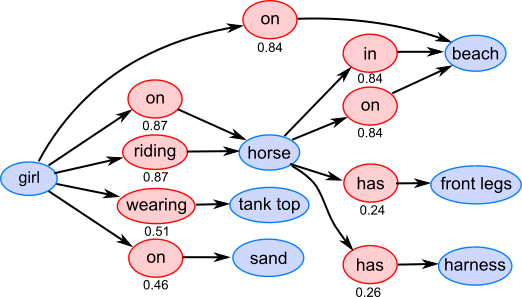}
\caption{Image input and scene graph output of our guided proposal framework. Object (blue nodes), predicate (class between two objects - red nodes) probabilities and a new relevance score for each object pair (below predicates) are computed using a CNN and Region-of-Interest (ROI) pooling on image crops.}
\label{fig:result-example}
\end{figure}

Furthermore, we present a novel relevance prediction scheme that evaluates how important a given object pair is to annotate. By focusing on the most relevant pairs and predicting several potentially correct predicates, as illustrated in Fig. \ref{fig:result-example}, our model is able to increase the diversity of predictions and thus to more fully exploit these dependencies.

\begin{figure*}[t]
\includegraphics[width=\linewidth, height=4cm]{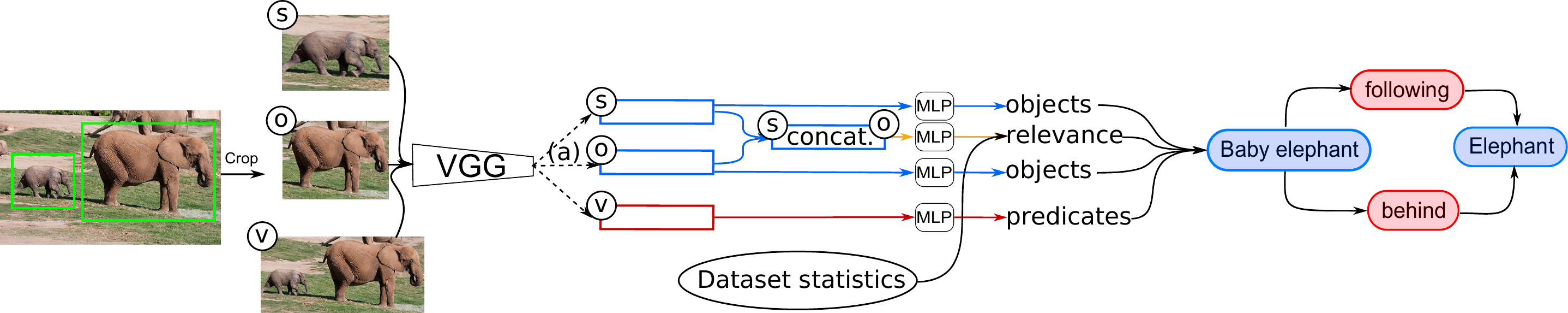}
\caption{Processing pipeline of our model at test time. (s), (o) and (v) refer to the subject, object and predicate of the relationship. (a) is the message passing operation introduced by \cite{Xu2017}. Feature vectors are passed to Multilayer perceptrons (MLP) to produce object and predicate class distributions, as well as to predict the relevance of the object pair (i.e. the probability of being annotated).}
\label{fig:model}
\end{figure*}

Work on such tasks has been enabled by the releases of large scale datasets providing bounding box annotations paired with natural language descriptions, or triplet annotations \cite{lu2016visual,krishnavisualgenome,Peyre_2017_ICCV}.
Many recent works have focused on learning to extract $<subject, predicate, object>$ triplets, and have shown that they overcome difficulties posed by the combinatorial nature of the problem, by using a language-vision multi-modal model \cite{lu2016visual}, exploiting semantic relationships between different triplets \cite{ramanathan_CVPR15}, statistical dependencies among the triplet constituents \cite{dai_CVPR_2017,li_CVPR_2017}, generating scene-graphs by producing object and relationship heatmaps \cite{Newell} or passing messages between object and predicate representations \cite{Xu2017}.

In \cite{galleguillos_CVPR_2008}, the authors show that integrating knowledge in the form of statistics measured on the studied dataset into the formulation of a Conditional Random Field makes outputs more consistent with the whole set of previous samples and thus increases its predictive power. Knowledge may also be integrated during the training phase of the model. Rohrbach et al. \cite{rohrbach_CVPR_2010} show that external knowledge on attributes allows zero-shot learning by associating classes with attributes and using common attributes to recognize instances of unseen classes. Specifically on visual relations detection, knowledge has been integrated in the form of semantic modeling of relations \cite{lu2016visual,ramanathan_CVPR15}.

In the same vein as \cite{galleguillos_CVPR_2008}, Yu et al. \cite{Yu_2017_ICCV} use predicate-object pair co-occurrences measured either on external (text corpora) or internal data (same dataset) to improve the consistency of the predictions. Contrary to \cite{galleguillos_CVPR_2008}, they integrate this knowledge during training using rule distillation, a process introduced by \cite{Hu_ACL_2016} to make a neural network learn to comply to diverse rules. It takes the form of a target distribution added to the loss function.


\section{Knowledge distillation}
A given predicate or object can be labeled in very diverse ways, especially when dealing with a large number of classes. To tackle this challenge, we integrate external knowledge into a network, in order to make better use of the statistical and semantic dependencies between object and predicate classes.

External knowledge may be integrated into neural networks using rule distillation as has been shown by \cite{Hu_ACL_2016} and more recently by \cite{Yu_2017_ICCV}, where internal and external knowledge are used to improve predicate classification.
However, one drawback of \cite{Yu_2017_ICCV} is that it is not possible to extract significant knowledge for each predicate when considering a context of several thousand classes, which we aim to tackle here. 
For this, we introduce a different semantic knowledge distillation scheme that is capable of treating a wider range of classes and increases scalability by limiting the burden of directly using large external corpora.

We define $\mathcal{C}$ and $\mathcal{V}$ as the sets of concept and predicate classes.
$s$ (subject concept) and $o$ (object concept) refer to instances of $\mathcal{C}$ and $v$ to instances of $\mathcal{V}$. $q$ and $p$ are used to refer to probability distributions.

Let us now consider a neural network with parameters $\theta$ that outputs a conditional probability distribution $p_{\theta}(\boldsymbol{Y}|\boldsymbol{X})$ of output variable $\boldsymbol{Y}$ given input variable $\boldsymbol{X}$.
As in \cite{Hu_ACL_2016,Yu_2017_ICCV}, we define $q(\boldsymbol{Y}|\boldsymbol{X})$ as
\begin{equation}
q = \mathrm{arg} \min_{q \in \mathcal{P}} \mathrm{KL}(q||p_{\theta}) - \lambda\mathbb{E}_q(f(\boldsymbol{X},\boldsymbol{Y})) \label{distill-eqn}
\end{equation}
where $q$ is the projection of $p_{\theta}$ on a subspace verifying constraints defined by $f$. The more $(\boldsymbol{X},\boldsymbol{Y})$ respects these constraints, the closer $f(\boldsymbol{X},\boldsymbol{Y})$ is to 1. $\mathrm{KL}(q||p_{\theta})$ is the Kullback-Leibler divergence from $p_{\theta}$ to $q$ and $\mathbb{E}_q(f(\boldsymbol{X},\boldsymbol{Y}))$ is the expectation of $f(\boldsymbol{X},\boldsymbol{Y})$ when the probability distribution of $\boldsymbol{Y}$ given $\boldsymbol{X}$ is $q(\boldsymbol{Y}|\boldsymbol{X})$. As shown in \cite{Hu_ACL_2016}, the closed form solution of Eq. \ref{distill-eqn} is $q(\boldsymbol{Y}|\boldsymbol{X}) \propto p(\boldsymbol{Y}|\boldsymbol{X})e^{\lambda f(\boldsymbol{X},\boldsymbol{Y})}$.

This new projected probability is added to the original network loss during the training:
\begin{align}
	L(\textbf{x},\textbf{y},\boldsymbol{\theta}) =  & (1-\pi^{(t)}) \cdot l(\boldsymbol{y}, p_{\theta}(\boldsymbol{Y}|\boldsymbol{x})) + \nonumber \\
    & \pi^{(t)} \cdot l(q(\boldsymbol{Y}|\boldsymbol{x}), p_{\theta}(\boldsymbol{Y}|\boldsymbol{x}))
\end{align}
where $l(\boldsymbol{y},p_{\theta}(\boldsymbol{Y}|\boldsymbol{x}))$ is the network loss, corresponding to the cross-entropy between the ground-truth label and the output distribution $p_{\theta}(\boldsymbol{Y}|\boldsymbol{x})$. $\pi^{(t)}$ is the weight of the distillation loss at iteration $t$. At the beginning of the training, since $p_{\theta}(\boldsymbol{y}|\boldsymbol{x})$ is far from the expected distribution, a large weight on the distillation loss would harm the training process, therefore $\pi^{(t)}$ is set close to 0 and increases during the training phase.

\subsection{Semantic knowledge distillation} \label{sk-distill}
Predicates are semantically similar when they appear in similar contexts. Hence in a given context, i.e. a given object pair, the probabilities of different predicates to describe this pair are related to their semantic similarity. We aim to use this knowledge by rewarding the model when semantically close predicates have similar probabilities. 

\begin{figure}[t]
\includegraphics[width=\linewidth]{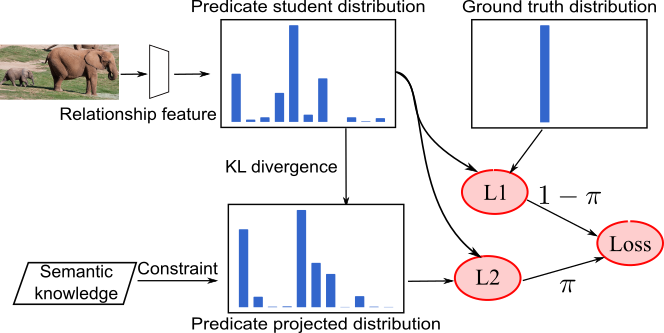}
\caption{Semantic knowledge is distilled into the network by projecting the output distribution under the constraint that predicates semantically similar to the selected predicate have high probabilities.}
\label{fig:semantic distill}
\end{figure}

Furthermore, several predicates may be true for a given object pair, thus the probability of one being true given that another is annotated is often greater than zero. For input $X=(I, b_1, b_2, (s,v,o))$, with $(s,v,o)$ the ground truth annotation for bounding boxes $b_1,b_2$ in image $I$ and output $Y=(s',v',o')$, we define
$f(X,Y) = \log(P(v'|v \in A_{r,I}))$ i.e. the probability of $v'$ being true for the pair $(b_1,b_2)$ given that predicate $v$ has been annotated. We model it by $P(v'|v \in A_{r,I}) \propto e^{\tau \cdot \mathrm{sim(v,v')}}$,
where $\mathrm{sim}(v,v')$ is the cosine similarity between embeddings of $v$ and $v'$. The embeddings are vector representations of words, computed such that words that frequently appear in the same context have embeddings with cosine similarity close to 1. $\tau$ is a temperature term which controls the entropy of the distribution. The lower $\tau$ is, the higher the entropy and the closer the distribution is to a uniform distribution. $\tau$ is set to 10,  allowing an object pair to have between 1 and 5 probable predicates (i.e. $P(v|v') \geq 0.1)$.\\
As illustrated in Fig. \ref{fig:semantic distill}, the projected distribution has increased probabilities for the predicates closest to the ground truth predicate and inversely for further predicates.\\
This formulation differs from the constraints expressed in \cite{Hu_ACL_2016} as we use the ground truth value to project the output distribution.
The loss gradient would be less stable if it was based on the output predicate instead of the ground truth, and this makes computations lighter as the constraints can be computed beforehand.

\subsection{Internal knowledge distillation} \label{sec:internal-knowledge}
We compare the previous distillation with internal knowledge distillation, presented by \cite{Yu_2017_ICCV}. 
The purpose of this method is to restrict the outputs to a subset of predicates that are the most probable for a given pair of objects.
\begin{equation} \label{internal-distill}
f(X,Y)=log(P(v|s,o))
\end{equation}
where $P(v|s,o)$ is computed on the training annotations.

Similarly to Section \ref{sk-distill}, the goal is to reward predicates frequently associated with the current context. For semantic knowledge distillation, this context was given by the annotated predicate. Besides, predicates were represented by vectors precomputed over a large text corpus, requiring fewer annotations but missing the specificities of image contexts. For internal distillation, the context is given by the object pair and since the predicate distribution is computed on the training annotation set, it is more accurate but requires more data to cover the whole space. To tackle the challenge of the long tail distribution of object classes, we regroup them by common words using a parser and a context free grammar to get the head word of a noun phrase.

\section{Guided relationship proposals}
With a high number of object detections, the number of pairs to annotate grows quadratically, which makes it important to select the most relevant pairs. For this, at test time, the model described in Fig. \ref{fig:model} is given a set of regions of interest and the goal is to correctly annotate a limited number of object pairs.
This setting differs from the training phase, where the model only learns to classify selected pairs of objects. Liang et al. \cite{liang_CVPR2017} recently showed that using a model that learns to choose pairs to annotate offers a much better predicting power.\\
In this section, we aim to improve the relationship scoring by prioritizing object pairs that are the most relevant, i.e. which are the most likely to be annotated by a human annotator and we show two complementary ways to achieve this goal.

We aim to model $P(r \in A_r|b_1,b_2,I)$: the probability with which the relationship $r=(s,v,o)$ will be annotated given the bounding boxes $b_1$ and $b_2$ and image $I$, with $A_{r,I}$ the set of relation annotations of $I$. This event is equivalent to the joint event: "$(s,v,o)$ is true" and "the $(s,o)$ object pair is annotated" which we note $(s,*,o) \in A_{r,I}$. For clarity, we omit the condition on $b_1,b_2,I$. 
\begin{align}
	P((s,v,o) \in A_{r,I}) &= P\left((s,v,o), (s,*,o) \in A_{r,I}\right) \nonumber \\
    & \approx P(s,v,o) \cdot P\left((s,*,o) \in A_{r,I}\right) \nonumber\\
    &= P(s,v,o) \cdot \mathrm{relevance}(s,o) \label{proposal-score} 
\end{align}
Here we make the hard assumption that the relevance of $(s,o)$ is independent from the predicate $v$. \\
This formulation departs from the usual formulation where only the first factor is considered when ranking proposals, which leads to less relevant results. 
		

\subsection{Relevance estimation} \label{data-driven-pair-scoring}
We first estimate relevance$(s,o)$ with statistics measured on the dataset:
\begin{equation}
\mathrm{relevance}(s,o) \approx \frac{n_{\mathrm{relations}}(s,o)}{n_{\mathrm{co-occurrences}}(s,o)}
		\end{equation}
For example: the number of relations between the classes "woman" and "ground" is very low when compared to the number of co-occurrences: this relationship is very common, thus humans tend not to prioritize "woman on ground". This object pair will then be penalized at test time.

A minimal value of 0.01 is used since this relevance matrix is very sparse.

\subsection{Relevance prediction} \label{sec:rel-pred}
To overcome the drawback of directly relying on object classes to estimate the relevance, we add a relevance prediction branch to the original network. It is based on Faster R-CNN \cite{Ren2015} and makes use of a message passing framework between object and predicate representations \cite{Xu2017}, as illustrated in Fig. \ref{fig:model}. It consists of three branches: predicate classification, object classification and bounding box regression. During the training phase, the loss corresponding to each branch is summed and back-propagated through the network.

Departing from the original architecture, we add an MLP for relevance prediction with two fully connected layers that take as input the feature vectors of both objects:
\begin{align} \label{relevance-pred-fc1}
\boldsymbol{x_{s,o}} = \boldsymbol{W_sx_s} + \boldsymbol{W_ox_o} + b_{s,o} \\
\mathrm{relevance(s,o)} \approx \boldsymbol{\sigma}(\boldsymbol{W_r x_{s,o}} + b)
\end{align}
where $\boldsymbol{\sigma}$ is the softmax function, $\boldsymbol{x_s}$ and $\boldsymbol{x_o}$ are the representations of the subject and object regions of interest and $\boldsymbol{W_s}, \boldsymbol{W_o}, \boldsymbol{W_{s,o}}, b_{s,o}, b$ are learnt weights and biases.
The corresponding loss is the binary cross-entropy between the output probability and the ground truth i.e. whether a given object pair has been annotated with a predicate, which is added to the unweighted global loss previously described.\\

\section{Experiments} \label{experiments}
We evaluate our model for proposing relationship triplets on three datasets varying in size and complexity. We compare it to the same model without knowledge distillation or relevance probability. 
We show that these contributions not only increase the overall performance of the original model but also give more diverse predictions which increases recall rates of rare predicates.

\subsection{Experimental settings}

\textbf{Datasets} \textbf{Visual Genome} (VG) \cite{krishnavisualgenome} consists of 108,077 images with object detections and predicate annotations for some object pairs. 
We remove object and predicate classes that appear only once in order to decrease the noise. We call this dataset Large VG. This version contains 20,000 object classes, 10,000 predicate classes and 1.8 million relationship annotations. In order to compare our method to existing ones, we also apply it on a filtered version of VG introduced by \cite{Xu2017}, restricted to 150 object classes and 50 predicate classes in 700,000 relationship annotations.
For both Large VG and Filtered VG, we use the training and test split defined by Xu et al. \cite{Xu2017}.

We also evaluate our method on \textbf{VRD} \cite{lu2016visual}, a dataset comprised of 4000 train images and 1000 test images annotated with 100 object classes and 70 predicate classes.

\textbf{Evaluation tasks and metrics} We evaluate our method on the following tasks defined in \cite{Xu2017}:
\begin{itemize}
\item \textbf{Predicate detection} (PredCls): ground truth object bounding boxes and classes are given and the model is evaluated on the quality of predicate prediction.
\item \textbf{Scene graph classification} (SGCls): only ground truth bounding boxes are given and we evaluate the quality of object and predicate classification.
\end{itemize}
To compare methods, we compute the R@k metric defined by the fraction of ground truth relationships retrieved among the top k predictions for a given image. As explained by \cite{lu2016visual}, the mean Average Precision metric is not used because it may penalize true predictions that do not appear in the ground truth annotations especially in the context which we consider, where many object and predicate classes may correctly describe a given pair of bounding boxes.

\textbf{Network implementation} To show the impact of our distillation approach, we use the network proposed by Xu et al. \cite{Xu2017}. This network architecture is based on the VGG-16 network \cite{Simonyan2015} pre-trained on MS-COCO \cite{LinMicrosoftContext}.
They add interconnected GRU cells with 512-dimension inputs and outputs to pass messages between object and relation nodes in order to refine predicted classes using context from the other objects and relationships. For comparison purposes we use the same layer configurations and hyper-parameters. \\
We also use the distillation hyper-parameters selected by \cite{Hu_ACL_2016} (i.e. $\lambda=6$ and $\pi(t)=\min(1-0.95^{\frac{t}{T}}, 0.1)$ where $t$ is the current iteration and $T$ is the maximum number of iterations. 

The word embeddings used by the semantic knowledge introduced in Section \ref{sk-distill} were obtained from the publicly available Glove model \cite{Pennington2014} trained on the Common Crawl corpus, consisting of 42B tokens.
\\

\begin{table}[t]
\begin{center}
\caption{Results on Large VG.} \label{tab:results-VG}
\begin{tabular}{l | c c | c c}
  \hline
  & \multicolumn{2}{c|}{PredCls} & \multicolumn{2}{c}{SGCls} \\
  & R@50 & R@100 & R@50 & R@100 \\
  \hline
  \hline
  Dual Graph \cite{Xu2017} & 22.65 & 32.69 & 8.58 & 11.15 \\
  \hline
  IK\cite{Yu_2017_ICCV}+ & 33.08 & 43.18 & 9.81 & 12.60 \\
  SK (Ours) & 33.33 & 43.39 & 9.84 & 12.57 \\
  SK - IK (Ours) &  33.04 & 43.04 & 9.93 & 12.73\\
  \hline
  $R_p$ (Ours) & 27.36 & 37.73 & 8.93 & 11.66 \\
  $R_e$ (Ours) & 42.08 & 51.54 & \underline{13.60} & \underline{16.93} \\
  $R_{p*e}$ (Ours) & 45.23 & \textbf{55.05} & \textbf{13.69} & \textbf{17.09}\\
  \hline
  IK-$R_e$ (Ours) & 45.13 & 54.67 & \underline{13.60} & \underline{16.93} \\
  SK-$R_e$ (Ours) & 45.14 & 54.67 & 13.36 & 16.71 \\
  SK-IK-$R_e$ (Ours) & \textbf{45.24} & \underline{54.74} & 13.59 & 16.89 \\
  SK-$R_{p*e}$ (Ours) & 44.93 & 54.37 & 13.48 & 16.76 \\
  \hline
\end{tabular}
\end{center}
\end{table}

\subsection{Results}

On the Filtered VG, we compare our results with the original network \cite{Xu2017} and the current state-of-the-art method of Newell and Deng \cite{Newell}. This last method is used to extract a scene graph in one pass over the image by producing heatmaps and predicting object and relationship properties at activated locations. Furthermore, we also report results for Dual Graph \cite{Xu2017}*, which are computed with at most one relationship proposal per object pair contrary to the other results where only the number of proposals per image is limited. 

\textbf{IK} stands for internal knowledge distillation and \textbf{SK} for semantic knowledge distillation. $\boldsymbol{R_e}$  denotes the data-driven relevance estimation (Section \ref{data-driven-pair-scoring}) and $\boldsymbol{R_p}$ the relevance prediction (Section \ref{sec:rel-pred}). Inspired by \cite{Hu_ACL_2016}, where the projected distributions are used at test time, we define $\boldsymbol{R_{p*e}} \triangleq * e^{\log(R_e)} = R_p*R_e$.

\textbf{Knowledge distillation}
On the Large VG dataset (Table \ref{tab:results-VG}), which constitutes the main focus of this work, and the main motivation for using the semantic distillation, both distillations bring significant improvements to the prediction task. A $10.5\%$ and $10.7\%$ increase of the R@100 over the original dual graph network is observed, corresponding to 32.1\% and 32.7\% relative gains.
Both distillations bring similar improvements overall, with internal distillation giving slightly better results on the filtered VG and VRD, and semantic distillation on Large VG. This shows the value of the presented semantic knowledge distillation, which incorporates knowledge from precomputed word representations into the neural network and can easily be applied to other benchmarks without requiring any additional data. 

Table \ref{tab:results-VRD} shows that on VRD, with a R@100 of 72.6\% for Dual Graph \cite{Xu2017} on the predicate classification task, the R@100 metric reaches 81.9\% with internal knowledge distillation and 80.8\% with semantic knowledge distillation. These results are outperformed by other methods presented by Yu et al. \cite{Yu_2017_ICCV} and Dai et al. \cite{dai_CVPR_2017} on the predicate classification task. However they make use of spatial features as input, which is outside the scope of this work.

\begin{table}[t]
\begin{center}
\caption{Results on Filtered VG.} \label{tab:results-filtered-VG}
\begin{tabular}{l | c c | c c}
  \hline
  & \multicolumn{2}{c|}{PredCls} & \multicolumn{2}{c}{SGCls} \\
  & R@50 & R@100 & R@50 & R@100 \\
  \hline
  \hline
  Dual Graph \cite{Xu2017}* & 44.75 & 53.08 & 21.72 & 24.38 \\
  Pixels to Graphs \cite{Newell} & \textbf{68.0} & 75.2 & 26.5 & 30.0 \\
  Dual Graph \cite{Xu2017} & 45.25 & 58.21 & 22.96 & 29.18 \\
  \hline
  $R_{p*e}$ (Ours) & 66.45 & 76.57 & 34.43 & 41.61\\
  IK-$R_e$  (Ours) & \underline{67.71} & \textbf{77.60} & \textbf{35.55} & \textbf{42.74} \\
  SK-$R_e$  (Ours) & 67.42 & \underline{77.43} & \underline{35.07} & \underline{42.25}\\
  \hline
\end{tabular}
\end{center}
\end{table}
\begin{table}[t]
\begin{center}
\caption{Results on VRD.} \label{tab:results-VRD}
\begin{tabular}{l | c c | c c }
  \hline
  & \multicolumn{2}{c|}{PredCls} & \multicolumn{2}{c}{SGCls} \\
  & R@50 & R@100 & R@50 & R@100 \\
  \hline
  \hline
  Region model \cite{Zhu} & 51.50 & 51.50 & \textit{N/A} & \textit{N/A} \\
  Dual Graph \cite{Xu2017} & 60.91 & 72.57 & 34.60 & 41.89 \\
  \hline
  IK (Ours) & \textbf{71.33} & \textbf{81.85} & \textbf{43.50} & \textbf{50.50} \\	
  SK (Ours) & \underline{71.02} & \underline{80.80} & \underline{41.19} & \underline{48.68} \\ 
  $R_e$ (Ours) & 66.27 & 76.81 & 39.18 & 46.50 \\
  $R_{p*e}$ (Ours) & 62.53 & 73.70 & 36.10 & 42.44\\
  SK-$R_e$ (Ours) & 69.19 & 79.35 & 41.89 & 48.82 \\ 
  \hline
\end{tabular}
\end{center}
\end{table}

\textbf{Guided relationship proposals}
Table \ref{tab:results-VG} shows a more significant improvement when weighting relationship proposals with the relevance score. With the relevance estimation, the recall is significantly increased, with a $57.7\%$ relative gain on the R@100, from $32.69\%$ to $51.54\%$. The difference in recall between $R_e$ and $R_p$ methods comes from the difference in scale in the two estimations: $R_e$ tends to have very low values except for a few pairs of classes which gives more opportunities to find the correct predicate, while $R_p$ is more homogeneous on a given image, thus increasing the number of pairs under consideration but also decreasing the recall for the most relevant ones. On the Large VG collection, the projected relevance prediction $R_{p*e}$ gives the best results, even greater than when combined with the semantic distillation. This highlights that the problem of correctly prioritizing object pairs plays a very important role in the capacity of the model to deal with use cases in which there very diverse classes.

The global recall does not capture the diversity of the predictions of the model as the VG is very unbalanced, with a few predicate classes making the vast majority of the annotations. We divide predicate classes by frequency in the Large VG: the top 10 most frequent predicates representing 78.3\% of the annotations, the next 30 with 10\% and the rest with the remaining 11.7\% annotations. In order to show the impact of our method on rarer predicates, we focus here on this second group and compute its macro R@100, i.e. we compute the R@100 for each predicate class of the group and average it. By incorporating the relevance estimation $R_e$ to the original network, the macro R@100 is multiplied by a factor of
11.7, increasing from 0.53\% to 6.9\%, which highlights that this problem is difficult and far from solved. For instance, in the case of predicates \textit{riding, hanging from, carrying} and \textit{parked on}, the recall increases from $2.3\%, 0.1\%, 0.4\%$ and $0.8\%$ to $50.6\%, 8.5\%, 12.3\%$ and $11.6\%$ respectively. For several predicates, the recall does not see a very important increase, which we assume comes from the fact that they are less context dependent and more dependent on the spatial configurations of objects, or that they are less visually meaningful (\textit{from, looking at etc.}). This is true for predicates \textit{in front of, at, over, from, covering} etc.

In Table \ref{tab:results-VRD} however, the relevance estimation gives a lower improvement than knowledge distillation, since there are fewer object annotations per image. On the Filtered VG (Table \ref{tab:results-filtered-VG}), the combination of internal knowledge distillation and relevance estimation reaches a new state of the art \cite{Newell}, with an improvement from $75.2\%$ to $77.6\%$ for predicate classification. Though the main purpose of this work was to improve predicate classification, we get a notable improvement of the  R@100 rate from $30\%$ to $42.74\%$ ($42.5\%$ gain) on the scene graph classification task. Semantic knowledge distillation provides the second best result by a small margin.

\section{Conclusion}
We proposed two complementary ways to incorporate knowledge and thus deal with some limitations of current visual relationship detection models. Firstly, by distilling external knowledge in a network we improve gradient stability during the training phase leading to a better global performance.
Secondly, by adding a relevance estimation at test time, either learnt or estimated on the dataset, we alleviate the problem of unbalanced classes and increase the diversity of the extracted scene graphs, thereby increasing the quality of the extraction.
Experiments on a two versions of the Visual Genome and on the VRD datasets show that either method brings significant improvements and that in the case of a large number of classes, the combination of the two is even more beneficial.
As a perspective, we aim to refine the prediction of both object pair relevance and predicate with the additional input of spatial features, and to propose a new knowledge distillation based on the statistics of these features.
The recent advances in reinforcement learning have, among others, made it possible for a model to learn to execute specific tasks, such as selecting bounding boxes to annotate and this constitutes another interesting perspective.

\small
\bibliographystyle{IEEEbib}
\bibliography{ICME}

\end{document}